\documentclass[sigconf,natbib=true]{acmart}
\AtBeginDocument{%
  }
\acmISBN{978-1-4503-XXXX-X/2018/06}
\setcopyright{acmlicensed}
\copyrightyear{2025}
\acmDOI{XXXXXXX}
\acmYear{2025}

\acmConference[SSTD'25]{19th International Symposium on Spatial and Temporal Data}{25-27 August, 2025}{Osaka, Japan}

\usepackage{cuted}
\usepackage{xcolor}
\usepackage{xspace}
\usepackage{caption}
\usepackage{balance}
\usepackage{amsmath} 
\usepackage{siunitx}
\usepackage{amsfonts}
\usepackage{tabularx}
\usepackage{multirow}
\usepackage{booktabs}
\usepackage{makecell}
\usepackage{hyperref}
\usepackage{subcaption}

\setcitestyle{numbers,sort&compress}

\newcommand{\pt}[1]{\left(#1\right)}
\newcommand{\psq}[1]{\left[#1\right]}
\newcommand{\pb}[1]{\left\lbrace#1\right\rbrace}
\newcommand{\E}[1]{\mathbb{E}\left[#1\right]} 

\begin{document} %%%%%%%%%%%%%%%%%%%%%%%%%%

\let\WriteBookmarks\relax
\def\floatpagepagefraction{1}
\def\textpagefraction{.001}

\title[Machine Learning-Driven Analysis of Global Port Significance and Network Dynamics]%
{
% $\textbf{\textit{Im}}\mathcal{PORT}\textbf{\textit{ance}}\mathcal{!}$
ImPORTance~--~Machine Learning-Driven Analysis of Global Port Significance and Network Dynamics for\\Improved Operational Efficiency
}

\author{Emanuele Carlini}
\email{emanuele.carlini@isti.cnr.it}
\orcid{0000-0003-3643-5404}
\affiliation{%
  \institution{Inst. of Info. Sci. and Technologies}
  \city{ISTI-CNR --- Pisa}
  \country{Italy}
}

\author{Domenico Di Gangi}
\email{digangidomenico@gmail.com}
\affiliation{%
  \institution{Inst. of Info. Sci. and Technologies}
  \city{ISTI-CNR --- Pisa}
  \country{Italy}
}

\author{Vinicius Monteiro de Lira}
\email{vinicius.monteiro@insightlab.ufc.br}
\orcid{0000-0002-7580-1756}
\affiliation{%
  \institution{Federal University of Ceará}
  \city{Fortaleza -- CE}
  \country{Brazil}
}

\author{Hanna Kavalionak}
\email{hanna.kavalionak@isti.cnr.it}
\orcid{0000-0002-8852-3062}
\affiliation{%
  \institution{Inst. of Info. Sci. and Technologies}
  \city{ISTI-CNR --- Pisa}
  \country{Italy}
}

\author{Amilcar Soares}
\email{amilcar.soares@lnu.se}
\orcid{0000-0001-5957-3805}
\affiliation{%
  \institution{Linnaeus University}
  \city{Växjö}
  \country{Sweden}
}

\author{Gabriel Spadon}
\authornote{Corresponding Author}
\email{spadon@dal.ca}
\orcid{0000-0001-8437-4349}
\affiliation{%
  \institution{Dalhousie University}
  \city{Halifax -- NS}
  \country{Canada}
}

\renewcommand{\shortauthors}{Carlini et al.}



% Keywords

\begin{abstract}
Seaports play a crucial role in the global economy, and researchers have sought to understand their significance through various studies. In this paper, we aim to explore the common characteristics shared by important ports by analyzing the network of connections formed by vessel movement among them. To accomplish this task, we adopt a bottom-up network construction approach that combines three years' worth of AIS (Automatic Identification System) data from around the world, constructing a Ports Network that represents the connections between different ports. Through this representation, we utilize machine learning to assess the relative significance of various port features. Our model examined such features and revealed that geographical characteristics and the port's depth are indicators of a port's importance to the Ports Network. Accordingly, this study employs a data-driven approach and utilizes machine learning to provide a comprehensive understanding of the factors contributing to the extent of ports. Our work aims to inform decision-making processes related to port development, resource allocation, and infrastructure planning within the industry.
\end{abstract}

\maketitle

\section{Introduction}

Seaports have garnered interest across economics, environmental studies, and social sciences. Understanding the role and significance of seaports (or simply \emph{ports}) is crucial for analyzing ocean mobility, identifying major hubs, and assessing the capabilities of these ports.

A standard method for assessing port relevance is analyzing vessel routes and port interconnections, often represented as a Ports Network, where ports are nodes and interconnections are edges. These edges can denote sequences of port visits by vessels~\cite{carlini2021understanding} or other maritime locations~\cite{varlamis2019network,varlamis2021building}. Modeling ports as networks allows the use of graph theory to analyze their roles and structure~\cite{ducruet2010ports, laxe2012maritime, alvarez2021maritime, del2021leveraging}, offering a topological perspective~\cite{zhang2023network}.
A central outcome of using network analysis on top of Ports Networks is understanding port centrality~\cite{cheung2020eigenvector,li2021sequence,wang2016determinants}, which reflects a port's topological importance.

As noted by Laxe et al.\cite{laxe2012maritime}, centrality captures a port's geographic significance within maritime routes. These measures include static properties ({\it e.g.}, degree centrality) and flow-based metrics ({\it e.g.}, Betweenness, Closeness, PageRank). Many studies link network centrality to port relevance~\cite{rodrigues2019network}, although few explore the shared traits among worldwide ports.
Historically, port network data came from shipping logs and insurance registries~\cite{ducruet2010ports, wang2016determinants}. More recently, Automatic Identification System (AIS) data have enabled detailed, large-scale Port Network modeling~\cite{yang2019big, wang2019extracting, carlini2021understanding, song2024gravity}, allowing precise vessel tracking and identification. AIS data also supports vessel traffic analysis and conservation planning for marine species~\cite{MCWHINNIE2021105479, spadon2024probabilistic, alam2024enhancing}.

In this paper, we construct a Ports Network from three years of global AIS data and examine how 34 features from the World Port Index\footnote{\url{https://msi.nga.mil/Publications/WPI}} can predict port centrality. Our goals are: (i) to assess how well port characteristics reflect their importance, and (ii) to identify standard features among central ports.
To this end, we train an AI-based model to predict port centrality using these features and analyze their importance with SAGE~\cite{covert2020understanding} and SHAP~\cite{lundberg2017unified}.

The feature importance analysis highlights that geographical location and port depth have the most influence in determining centrality. Building on these findings, our contributions are:
\begin{itemize}
    \item The design of a centrality measure that combines multiple definitions of centrality used in literature and its application to a Ports Network built with three years of AIS data;
    \item The application and evaluation of a machine learning classification task to approximate port centrality in the network from the features extracted from their structure; and,
    \item An extensive feature importance analysis to evaluate relevant features of ports in estimating centrality.
\end{itemize}

In order to present these contributions, we structured this paper as follows: Section~\ref{sec:related} reviews prior work on port centrality. Section~\ref{sec:methodology} details the dataset, Ports Network construction, centrality definition, and machine learning methods. Section~\ref{sec:evaluation} presents key findings and feature importance results. Section~\ref{sec:conclusion} concludes the paper.

\section{Related Works}
\label{sec:related}

We analyzed and compared the most relevant works in literature along three dimensions stated below (see Table \ref{tab:related_work} for more details):
\begin{itemize}
    \item \textit{Source data}. Ports Networks are usually built from two types of datasets: \textbf{(i)} bottom-up approaches use fine-grained data, such as AIS data; here, the connection between ports and vessel routes are extracted with an extensive data processing, such as in \cite{yang2019big}; \textbf{(ii)} top-down approaches use ship schedules and historical registries of vessels routes to build the network. \emph{We used three years (2017-2019) to build a Ports Network.}
    
    \item \textit{Network scope}. Studies vary regarding the location of the ports. Analysis can be focused on a specific area or country, such as the Canary Islands \cite{tovar2015container}, or global trends ({\it i.e.}, worldwide and continental). \emph{We have analyzed all the world's~ports.}
    
    \item \textit{Centrality measures}. A wide variety of centrality measures have been used in the literature. Our analysis derives an aggregated centrality measure combining the Degree, PageRank, Betweenness, and Closeness Centralities. Proposing a new technique focused on the importance of nodes, and with a specific aim of assessing ports in such a Port Network.
\end{itemize}

\begin{table}[!htbp]
    \centering
    \footnotesize
    \begin{tabular}{c l l c}
         & \textbf{Data Source (Period)} & \textbf{Network Scope} & \textbf{Centrality}\\
         \midrule
         \cite{ducruet2010centrality} & LSI (1996-2006) & Local (NE Asia) & MD, B\\
         \midrule
         \cite{laxe2012maritime} & LSI (2008-2010) & Local (China) & B\\
         \midrule
         \cite{montes2012general} & LSI (2008-2010) & Global (World) & B\\
         \midrule
         \cite{seoane2013foreland} & AIS (Mar 2007-08, 2010-11) & Local (Europe) & B, D\\
         \midrule
         \cite{tran2014empirical} & CIY (1995-2011) & Global (E-W lane) & D*, WD*\\
         \midrule
         \cite{tovar2015container} & SLS (Oct 2012) & Local (Canary Island) & D, B, AI\\
         \midrule
         \cite{kosowska2016evolving} & LSI (1890-2000) & Global (World) & C, B\\
         \midrule
         \cite{wang2019extracting} & AIS (2015) & Global (World) & D, B\\
         \midrule
         \cite{cheung2020eigenvector} & SLS (Q4 2015) & Global (World) & E\\
         \midrule
         \cite{tocchi2022hypergraph} & SLS (2019) & Global (World) & B*, D\\
         \midrule
         \cite{wang2016determinants} & SLS (N/S) & Global (39 ports) & D, C, B\\
         \midrule
         \cite{li2021sequence} & AIS (2017) & Regional (Chesapeake Bay) & PR, PC\\
         \midrule
         \textit{Ours} & AIS (2017-2019) & Global (World) & D, PR, B, C\\
         \bottomrule
    \end{tabular}
    \caption{Comparison of relevant works in centrality measurement on port networks --- the asterisk ``*'' indicates an adapted version of the centrality with respect to the original formulation in classical graph theory;
    \textbf{\textit{Acronyms:}} LSI -- Lloyd's Shipping Index, AIS -- Automatic Identification System, CIY -- Containerisation International Yearbooks, SLS -- Shipping lines schedule, MD -- Maritime Degree, B -- Betweenness, D -- Degree, WD -- Weighted Degree, C -- Closeness, E -- Eigenvector, PR -- PageRank, PC -- Participation Coefficient, AI -- Accessibility Index, N/S -- Not Specified.}
    \label{tab:related_work}
\end{table}

Ducruet et al.~\cite{ducruet2010centrality} are among the first to systematically study cargo networks with complex network techniques at a large scale. In one of their first papers~\cite{ducruet2010centrality}, they provide an empirical analysis of the centrality of ports in Northeast Asia by using inter-port traffic flows from the Lloyd's Shipping Index\footnote{The insurance company Lloyd's has historically collected movement of vessels between two or more ports daily or weekly since 1890. Their data includes the dates of departure and arrival, tonnage capacity, operating company, flag, and additional information on the voyage. Nowadays, Lloyd's covers about 80\% of the world's fleet.}.
The work of~\cite{laxe2012maritime} also utilizes a sample of the Lloyd's dataset, which includes the movements of the world's container ship fleet from Chinese ports from 2008 to 2010. 
Their work aims to examine the maritime networks before and after the 2008 financial crisis, analyzing the extent to which large ports have seen their position within the network change. The authors demonstrate how the global and local significance of ports can be quantified using concepts from graphs.

A similar analysis is conducted by~\cite{montes2012general}, which uses the same centrality metrics to observe the evolution of the Ports Network of container vessels on a broader geographical scale for 2 months of 3 consecutive years.
A complete historical analysis of the Lloyds dataset was conducted by \cite{kosowska2016evolving}. 
They investigate the structure of the maritime trade network and examine its relationship to efficiency.
Given the extended coverage of their data (1890-2000), the authors assessed the topological over-time change of the corresponding network, which evolved from a highly clustered network towards a hub-and-spoke network.
Cheung et al.~\cite{cheung2020eigenvector} utilized eigenvector centrality as a decision-making tool for predicting potential new links between ports that would enhance network connectivity. In other words, the eigenvector centrality becomes the objective of a max-min optimization problem. They built the network using information from major shipping lines in the fourth quarter of 2015, resulting in a graph comprising 601 nodes (ports) and 3,737 links.

The work of~\cite{seoane2013foreland} employs complex network techniques to analyze the growth rates of European ports. The ``proximal foreland'' (essentially a subgraph of a Ports Network considering all ports at three hops of a given center port) is used to measure the connectivity of the ports. They analyzed the AIS data of one month of four different years (2007, 2008, 2010, 2011) to extrapolate cargo vessel trajectories.
AIS data were also utilized by~\cite{wang2019extracting}, who constructed a Ports Network using the 2015 worldwide AIS data with multiple spatial levels. Their bottom-up process mainly consists of five steps, where the first three generate the network nodes, and the last two create the links. They apply the DBSCAN (Density-Based Spatial Clustering of Applications with Noise) algorithm to detect where ships stop and cross, and then combine this information with terminal candidates of ports. A directed Port Network is generated with the trip statistics between two nodes as the edges. They evaluate features such as the average Degree and Betweenness node centrality, the average shortest path length between any two nodes, and community clusters within their proposed Ports Network.

Several works employ a modified version of the original centrality metrics to adapt for the peculiarity of Port Networks.
The work presented in~\cite{tran2014empirical} analyses the East-West lane over several years. They compute a set of indicators to evaluate the evolution of connectivity in terms of degree centrality, which is also calculated by considering the amount of TEU (twenty-foot equivalent units) exchanged between ports.
The work of \cite{tocchi2022hypergraph} focuses on applying centrality measures on a hypergraph network of container services. They provide centrality for (hyper) P-graphs, which represent direct port-to-port service connections, and (hyper) L-graphs, where the edges represent the transit of a container vessel between two ports. For reference, our paper considers the network as an L-graph.
The Betweenness Centrality for hypergraphs is computed using the probability that the shortest path goes through a specific node in the hypergraph (rather than the yes/no formulation of the classical Betweenness Centrality).
In \cite{wang2016determinants}, the authors define the importance of a port by applying a \textit{centrality index}, composed of Freeman's measures of Degree, Closeness, and Betweenness Centrality measures. They utilize these centrality measures to select 39 container ports worldwide (no information about the temporal range provided).

Network analysis has also been used to explore the relevance of ports localized in specific regions. \cite{tovar2015container} assesses the connectivity of the main Canarian ports with various centrality measures ({\it e.g.} Degree, Betweenness, and Port Accessibility Index). They generated a network of 53 ports directly related to Las Palmas and Tenerife ports, using one month (October 2012) of the shipping line schedule.
More recently, the work of \cite{li2021sequence} used AIS data to generate trajectory sequences and assess the importance of way-points in the Chesapeake Bay area. They employed one year (2017) of AIS data to generate a fine-grained network of an enclosed area and compute a centrality analysis of way-points. To assess the importance of way-points, they use both centrality measures (PageRank) and community-based measures, {\it i.e.}, the participation coefficient (the strength of a node's connections within its community).

\begin{strip}
    \begin{minipage}{\textwidth}
        \centering
        \includegraphics[width=0.99\textwidth]{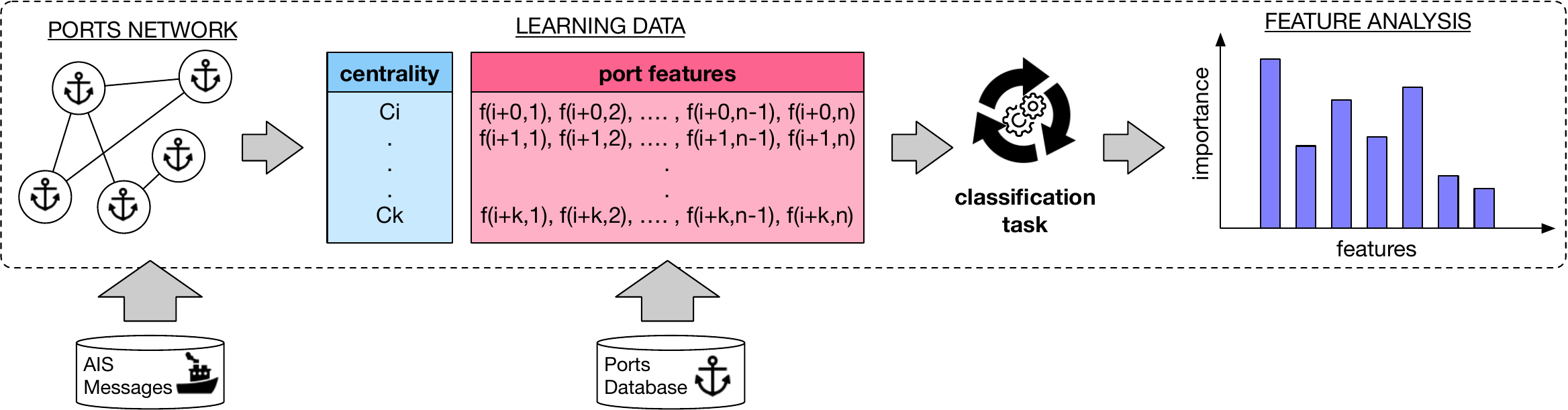}
        \captionof{figure}{Methodological Framework for Port Centrality and Feature Analysis. This diagram outlines the methodology used to analyze port centrality and its features following the bottom-up approach to creating our Ports Network. Starting from the AIS messages and port databases, the Ports Network is constructed to calculate centrality measures. Because various features form the basis of the data used for the classification task, the feature analysis identifies key factors contributing to port centrality.}
        \label{fig:methodology}
    \end{minipage}
\end{strip}

\section{Dataset \& Methodology}
\label{sec:methodology}

This paper investigates how port features relate to their importance in maritime networks of ports interconnectivity.
Through our methodology we seek response for two queries:
\textbf{(RQ1)} \textit{How effective is it to assess port importance by solely considering its features?} and \textbf{(RQ2)} \textit{What are the common features found in the most central ports?}

We implemented the methodology in Figure \ref{fig:methodology} to answer these questions. 
The Ports Network is built by using two datasets: \textbf{(i)} a publicly available dataset containing the port features and position, and \textbf{(ii)} an AIS messages dataset that contains information about vessels for the period 2017-2019.
The importance of ports is then defined by considering their centrality in the network. 
The relationship between importance and features is assessed using a machine learning model trained to predict significance based on features.

\subsection{Data Sources}
\label{sec:data}

\noindent
\textbf{AIS data}. It is a key data source for tracking vessel movements, transmitting positions and activity data globally
Each message contains several pieces of information regarding a vessel, including the ship identifiers, coordinates, and other data such as kinematics and vessel type. More details on the format and applications of AIS messages can be found in~\cite{yang2019big}.
In this work, we used three years (2017-2019) of worldwide AIS data provided by Spire\footnote{\url{spire.com} (now operated by \url{kpler.com})}.
The raw dataset is private and contains 20 billion messages, around 2.5 TB.

\noindent
\textbf{Port data}. 
Ports' features are taken from the 2020 World Port Index dataset\footnote{\url{msi.nga.mil/Publications/WPI}} (WPI), a publicly available dataset released by the National Geospatial-Intelligence Agency of United States.
The dataset contains information on about 3,630 ports worldwide. In total, 79 features are reported for each port. They include various port identifiers, positions, port characteristics, facilities, and available services.

\subsection{Ports Network}

Vessels report their positions ({\it i.e.}, coordinates) via AIS messages during navigation. We processed these messages to capture the fine-grained spatio-temporal trajectories of individual vessels. From these, we extracted sequences of port visits for each unique vessel, ordered by time.
All vessel sequences were then merged to construct a unified Ports Network, where nodes represent ports and edges capture traffic between them. An edge is created when a vessel visits two ports consecutively, and its weight reflects the total number of vessels traveling between those ports, regardless of direction.
Construction the Ports Network involves three steps:
\textbf{(i)} extracting port visits from raw AIS data;
\textbf{(ii)} aggregating port-to-port transitions; and
\textbf{(iii)} generating the network from the compiled visit sequences.
These procedures are detailed in our earlier work~\cite{carlini2020uncovering, carlini2021understanding}.

\subsubsection{Port Visits}

This phase aims to identify vessel visits to ports. First, we filter AIS messages transmitted within defined port areas -- modeled as circular buffers centered on each port's coordinates. To handle overlapping buffers ({\it i.e.}, AIS messages that fall within multiple ports), we merge overlapping areas and retain the largest port, determined using the HARBORSIZE attribute.
Vessels are uniquely identified by a combination of MMSI, IMO number, and call sign; entries lacking this information were excluded.
Next, for each vessel, AIS messages are chronologically sorted to extract a sequence of port visits. Consecutive messages from the same port are collapsed into a single visit. This process yielded approximately 4.5 million visits by nearly 100,000 uniquely identified vessels.

\subsubsection{Generating the Ports Network}
\label{subsec:graph_generation}

We constructed the Ports Network using only \textit{cargo} vessels, totaling approximately 1.5 million port visits from 30,000 unique vessels. From these visits, we derived each vessel's sequence of voyages, where a voyage is defined as the movement from an origin port $p_o$ to a destination port $p_d$ based on consecutive timestamps $t_i$ and $t_{i+1}$.
This data was used to build a directed graph $G = {V, E}$, where $V$ denotes ports and $E$ represents directed edges corresponding to vessel movements. Initially, this produced a multi-graph with one edge per voyage. We then collapsed multi-edges into single directed edges weighted by the frequency of voyages between port pairs.
The resulting network consists of multiple strongly connected components. Our analysis focuses on the largest one, which includes 1,154 nodes (93\%) and 21,776 edges (99\%), of which 7,544 (35\%) are bi-directional. The network is sparse, with a density of 0.01, a diameter of 10, an average shortest path length of 3.1, and a clustering coefficient of 0.6.

\subsection{Port Centrality}

Several centrality measures can be utilized to the assessment of the relevance of a port.
To be as general as possible, we have chosen to measure centrality by combining the most used metrics in the literature we reviewed.
Specifically, we define port centrality as the function $A(\cdot)$ defined for each port $p \in V$ in the Ports Network. $A(\cdot)$ compute the aggregation of the following centrality measures:

\begin{itemize}
    \item \textit{InDegree} (DI) and \textit{OutDegree} (DO) are defined by counting, respectively, the number of incoming and outgoing links for each node. In this context, the DO of a port refers to the number of ports from which vessels depart, while the DI counts the number of ports to which vessels arrive.
    
    \item \textit{PageRank} (PR) and its weighted variant (wPR) are widely used algorithms for evaluating node importance in complex networks. Initially developed to rank web pages, PR has since been applied across various domains, including social networks and biological systems. The algorithm assigns importance to a node based on the significance of its neighbors. In our context, a port's relevance is proportional to the importance of other ports connected to it via vessel movements. In the classical (unweighted) PR, each neighboring port contributes equally to the overall score. In contrast, the weighted version (wPR) adjusts contributions based on the logarithm of the number of trips between connected ports.
    
    \item \textit{Betweenness Centrality} (BC). The BC is a topological measure of a node's importance. It is widely adopted for analyzing a wide range of complex networks, particularly those with a concept of information flow between nodes (in a Ports Network, the flow represents the traffic between ports). The BC works by counting the number of times a node is in the shortest path between all node pairs in the network, as in:
    \begin{equation}
        \centering
        BC(p) = \sum_{p \neq s \neq d}{\frac{\sigma_{sd}(p)}{\sigma_{sd}}}
    \end{equation}
    where $\sigma_{sd}$ is the number of shortest paths from a starting node $s$ and a destination node $d$, and $\sigma_{sd}(p)$ is the number of shortest paths that go through $p$.
    In our case, the flow is represented by the maritime traffic between ports, and the importance of a port is proportional to the presence of the port in the critical routes of the global Ports Network. 
    
    \item \textit{Closeness Centrality} (CC). It measures the closeness of a node to all others.
    The CC of a node $n$ is the reciprocal of the average shortest path distance from it to all reachable ones:
    \begin{equation}
        \centering
        CC(n) = \frac{|N|}{\sum_{y \in N} d(y, n)}
    \end{equation}
    where $N$ is the set of all nodes in the network; indicating the proximity of a port in terms of "hops" to other ports.
\end{itemize}

Applying each of those measures to the Ports Network produces one centrality measurement for each node. Therefore, for each port $p$ we compute the $6$ centralities described above resulting in:
\begin{equation}
    \mathscr{C} = \{ DI(p), DO(p), PR(p), wPR(p), BC(p), CC(p)\}.
\end{equation}

To combine the different measures, we first standardize each by computing the z-score, which involves subtracting the mean and then dividing by the standard deviation. In the case of a generic centrality $c \in \mathscr{C}$ for a port $p$ and $N$ ports, formalized as: 
\begin{equation}
    \centering
    z(c, p) = \frac{c(p) - \frac{\sum_p c(p)}{N}}{ \sqrt{\sum_p c(p)^2 - \left(\sum_p c(p)\right)^2}}
\end{equation}

Finally, the final measure of the aggregated centrality of a port is the average of the standardized centralities, formalized as:
\begin{equation}
    \centering
    \label{ed:centrality}
    A(p) = \frac{\sum_{c \in \mathscr{C}} z(c,p)}{\lvert \mathscr{C} \rvert}
\end{equation}

\subsection{Port Features}
\label{sec:port_features}

\begin{table*}[h!]
\resizebox{.85\textwidth}{!}{
\footnotesize
\begin{tabular}{llccccc}
\toprule
   \textbf{Category}  &   \textbf{Attribute}      &  \textbf{Missing(\%)} &  \textbf{Support} &  \textbf{Cardinality} &  \textbf{Min Distr. (\%)} &  \textbf{Max Distr. (\%)} \\
%Category & Attribute &             &          &              &               &               \\
\midrule
\multirow{6}{*}{COMMUNICATION} & \textbf{AIR} &        41.3 &     2130 &            2 &           3.4 &          96.6 \\
     & FAX &        66.0 &     1234 &            2 &           2.6 &          97.4 \\
     & \textbf{PHONE} &        47.2 &     1915 &            2 &           3.1 &          96.9 \\
     & \textbf{RADIO} &        29.6 &     2556 &            2 &           1.1 &          98.9 \\
     & RADIO TEL &        56.9 &     1563 &            2 &           3.0 &          97.0 \\
     & RAIL &        50.6 &     1795 &            2 &           2.4 &          97.6 \\
\cline{1-7}
\multirow{3}{*}{CRANE} & FIXED &        61.8 &     1387 &            2 &           6.9 &          93.1 \\
     & FLOAT &        82.0 &      652 &            2 &          16.3 &          83.7 \\
     & MOBIL &        54.3 &     1658 &            2 &           5.2 &          94.8 \\
\cline{1-7}
\multirow{5}{*}{DEPTH} & \textbf{ANCHORAGE} &         9.8 &     3274 &           16 &           0.1 &          18.5 \\
     & \textbf{CARGO DEPTH} &        12.5 &     3175 &           16 &           0.3 &          16.2 \\
     & \textbf{CARGO WHARF} &        24.3 &     2747 &            2 &           0.3 &          99.7 \\
     & \textbf{CHANNEL} &        12.6 &     3171 &           16 &           0.9 &          11.5 \\
     & OIL WHARF &         0.0 &     1689 &           15 &           1.5 &          14.7 \\
\cline{1-7}
\multirow{4}{*}{ENTRANCE RESTRICTIONS} & \textbf{ICE} &        26.1 &     2681 &            2 &          22.9 &          77.1 \\
     & \textbf{OTHER} &        15.6 &     3064 &            2 &           7.9 &          92.1 \\
     & \textbf{SWELL} &        22.1 &     2827 &            2 &          27.1 &          72.9 \\
     & \textbf{TIDE} &        24.6 &     2736 &            2 &          29.2 &          70.8 \\
\cline{1-7}
LIFT & 50-100 TONS &        79.5 &      743 &            2 &           7.7 &          92.3 \\
\cline{1-7}
\multirow{18}{*}{OTHER} & CARGO\_ANCH &        56.7 &     1573 &            2 &           2.3 &          97.7 \\
     & \textbf{DIRTY BALLAST} &        28.4 &     2600 &            2 &          34.7 &          65.3 \\
     & DRYDOCK &        79.4 &      748 &            3 &          19.0 &          43.9 \\
     & \textbf{ETA MESSAGE} &        15.9 &     3054 &            2 &          10.8 &          89.2 \\
     & \textbf{FIRST PORT OF ENTRY} &        37.8 &     2259 &            2 &          24.3 &          75.7 \\
     & GARBAGE DISPOSAL &        55.7 &     1608 &            2 &          20.9 &          79.1 \\
     & \textbf{HARBOR SIZE} &         0.2 &     3624 &            4 &           4.4 &          58.6 \\
     & \textbf{HARBOR TYPE} &         0.0 &     3618 &            8 &           0.9 &          34.9 \\
     & \textbf{HOLDGROUND} &        47.7 &     1899 &            2 &          13.7 &          86.3 \\
     & \textbf{MAX SIZE VESSEL} &        19.5 &     2921 &            2 &          39.1 &          60.9 \\
     & \textbf{MEDICAL FACILITIES} &        23.5 &     2776 &            2 &           3.2 &          96.8 \\
     & \textbf{OVERHEAD LIMITATION} &        48.5 &     1868 &            2 &          39.2 &          60.8 \\
     & RAILWAY &        58.8 &     1497 &            3 &           9.6 &          65.4 \\
     & \textbf{REPAIR\_CODE} &        24.6 &     2737 &            5 &           5.2 &          60.7 \\
     & \textbf{SHELTER} &         0.9 &     3599 &            5 &           1.0 &          35.7 \\
     & \textbf{TIDE\_RANGE} &         0.0 &     3587 &           14 &           0.5 &          32.4 \\
     & TURNING AREA &        63.3 &     1332 &            2 &          11.4 &          88.6 \\
     & \textbf{US REPRESENTATIVE} &        38.2 &     2244 &            2 &          14.4 &          85.6 \\
\cline{1-7}
\multirow{3}{*}{PILOTAGE} & ADVISABLE &        64.5 &     1290 &            2 &           5.3 &          94.7 \\
     & \textbf{AVAILABLE} &        30.5 &     2522 &            2 &           5.1 &          94.9 \\
     & \textbf{REQD} &        19.9 &     2907 &            2 &          14.6 &          85.4 \\
\cline{1-7}
\multirow{2}{*}{QUARENTINE} & OTHER &        69.2 &     1119 &            2 &           0.3 &          99.7 \\
     & \textbf{PRATIQUE} &        45.0 &     1998 &            2 &           2.1 &          97.9 \\
\cline{1-7}
\multirow{2}{*}{SERVICES} & ELECTRICAL &        72.5 &      999 &            2 &          12.1 &          87.9 \\
     & \textbf{LONGSHORE} &        46.4 &     1946 &            2 &           5.8 &          94.2 \\
\cline{1-7}
\multirow{5}{*}{SPATIAL} & \textbf{COUNTRY} &         0.0 &     3029 &           48 &           0.5 &          22.0 \\
     & LAT\_HEMISPHERE &         0.0 &     3629 &            2 &          13.8 &          86.2 \\
     & LONG\_HEMISPHERE &         0.0 &     3629 &            2 &          47.3 &          52.7 \\
     & \textbf{LATITUDE} &         0.0 &     3629 &            cont. &          n/a &          n/a \\
     & \textbf{LONGITUDE} &         0.0 &     3629 &            cont. &          n/a &          n/a \\
\cline{1-7}
\multirow{6}{*}{SUPPLIES} & DECK &        63.8 &     1315 &            2 &          24.0 &          76.0 \\
     & \textbf{DIESEL} &        38.6 &     2228 &            2 &          15.9 &          84.1 \\
     & ENGINE &        63.7 &     1316 &            2 &          23.0 &          77.0 \\
     & \textbf{FUEL OIL} &        26.0 &     2686 &            2 &          14.7 &          85.3 \\
     & \textbf{PROVISIONS} &        37.9 &     2255 &            2 &           7.2 &          92.8 \\
     & \textbf{WATER} &        13.6 &     3136 &            2 &           6.6 &          93.4 \\
\cline{1-7}
\multirow{2}{*}{TUGS} & \textbf{ASSIST} &        26.2 &     2680 &            2 &          22.2 &          77.8 \\
     & SALVAGE &        67.7 &     1172 &            2 &          23.5 &          76.5 \\
\bottomrule
\end{tabular}
}
\caption{Summary of Key Port Features in the WPI Dataset. Features are grouped by category. Features retained after the cleaning phase are shown in \textbf{bold}. Missing indicates the percentage of null. Support is the number of ports with valid entries. Cardinality is the number of unique feature classes. Min and Max is the proportion of the least and most frequent classes.}
\label{tab:port_features}
\end{table*}

The WPI dataset (see Section \ref{sec:data}) contains $70+$ features for 3,630 ports. The preprocessing of the WPI dataset included cleaning operations and filling empty data. 
From the entire dataset, we removed those features that represent meta or external references, such as the region number and the port name. From the remaining features, as listed in Table \ref{tab:port_features}, we removed those with a percentage of missing values exceeding 50\%. This results in $36$ features (in \textbf{bold} in Table \ref{tab:port_features}) for the analysis, of which $34$ are categorical and $2$ are continuous.

We then impute the remaining missing values using a multivariate iterative approach. That amounts to sequentially fitting a regression to explain the non-missing values of one column as a function of all the other columns as discussed, for example, in \cite{liu2013comparison}. The regression is used to infer the missing values of the column, and this procedure is repeated for each column. The whole cycle of the regression fit and imputations is repeated $10$ times\footnote{We used scikit-learn's iterative imputation: \url{tinyurl.com/35b399hn}.}.

\subsection{Predicting Ports Relevance}
\label{sec:ml}

To predict relevance from the features, we first consider a binary classifier to determine whether a port is among the most central ones. We experimented with thresholds to separate relevant and non-relevant ports, considering the top $5\%$, $10\%$, and $15\%$ as relevant.
We use $75\%$ of the dataset for the training and $25\%$ for evaluation.

We examined various machine learning models for the binary classification task and consistently obtained similar results across different models.
We selected Random Forest to distinguish central from non-central ports effectively.
The following section discusses the results obtained from this approach, which mitigates the high variance found in single trees by averaging their numbers.
Each tree grows considering only a random subset of features before each split. 
After a fixed number of different trees are grown in this way, their predictions are combined with a majority vote.
We used $100$ trees, grown using Gini impurity to guide tree splits.

\subsection{Evaluate Feature Importance}

Explaining the outcomes of machine learning models is a significant problem in many contexts and has garnered considerable attention in the literature. Various interpretability tools and approaches have been proposed to improve our understanding of complex models \cite{lundberg2017unified, ribeiro2016should, sundararajan2017axiomatic}. Here we consider two approaches, SHapley Additive exPlanations (SHAP) \cite{lundberg2017unified} and Shapley Additive Global importancE (SAGE) \cite{covert2020understanding}, that quantify how much a model relies on each feature in making predictions and define feature importance as the amount of predictive power that can be associated to it. While both methods are inspired by game theory concepts and are based on Shapley values, they differ in the scope of the feature importance they define.
On the one hand, SHAP explains the outcome of individual predictions, thus focusing on local interpretability. On the other hand, SAGE describes the model's behavior across the entire dataset, which we refer to as global interpretability. In the following, we provide an overview of the two methods and refer the interested reader to the original papers for more detailed information.

SHAP and SAGE are based on the Shapley values~\cite{shapley1953value}. These methods enable the computation of each feature's importance while considering its interactions. To do so, they repeatedly evaluate the model with various input feature sets derived from the original one by ``soft-removing" the information content of features not included in the set. The removal of a feature is achieved by shuffling its values, thereby destroying any potential contribution to the prediction.
In the following, we indicate random variables with capital letters and observations in lowercase. Given the response variable $Y$, the set of features $X$, consisting of individual features $X_1, \dots, X_d$, a subset $S$ of the features is denoted by $X_S$.
Considering a model $f\pt{X}$, the methods consider a feature necessary if its absence harms the model outcome, where the model impact is quantified differently in SHAP and SAGE. Instead of considering the removal of a single feature, both approaches, based on Shapley values, compare all possible subsets of features, including or excluding each feature.

\subsubsection{Local Interpretability}

SHAP~\cite{lundberg2017unified} captures the relevance of each feature for an individual prediction, rather than considering each feature's relevance for the entire set of predictions. In our application, SHAP answers questions like \textit{how relevant was feature cargo depth in classifying Port Keppel as central (non-central)?}

SHAP starts from the prediction for one observation $x$\footnote{In our case, each observation is a port with its set of features.}, based features $x_S$. That is defined as $v_{f, x}\pt{S} = \E{f\pt{X} \vert X_S = x_S}$ where the expectation is taken over the rest of the features not in $S$. Those features are marginalized out, and their contribution is ``softly-removed". From this the SHAP value for $x$ $\phi_i\pt{f, x}$ is defined as:
\begin{equation}
    \phi_i\pt{f, x} = \frac{1}{d} \sum_{S\in D \setminus i} \binom{d-1}{|S|} \psq{v_{f, x}\pt{S \cup \pb{i}} - v_{f, x}\pt{S}}
\end{equation}
A negative value indicates that including a feature in the model decreases the expected prediction for the observation.

\subsubsection{Global Interpretability}
SAGE focuses on the global importance of a feature for all observations available. This amounts to assigning a number to each feature to quantify its relevance for predicting all observations. In our application, SAGE answers questions like \textit{how relevant was feature cargo depth in classifying ports as central or non-central?}
They consider a loss function $l\pt{f\pt{X}, Y}$, {\it i.e.}, the cross entropy, and define $f_S\pt{x_S} = \E{f\pt{X} \vert X_S = x_S}$ where the expectation is taken over the rest of the features not in $S$. Then they define the amount of predictive power that a model derives from the set of features $S$ as $v_f\pt{S} = \E{l\pt{f_{\emptyset}\pt{X_\emptyset}}} - \E{l\pt{f_{S}\pt{X_S}}}$. To obtain the importance of individual feature $i$ from $v_S$ (a set $S$ of possibly multiple features), they compare all the sets, including or not including feature $i$ -- the SAGE value $\phi_i$ is:
\begin{equation}
    \phi_i = \frac{1}{d} \sum_{S\in D \bmod i} \binom{d-1}{|S|} \psq{v\pt{S \cup \pb{i}} - v\pt{S} }
\end{equation}

% Note that, differently from SHAP, 
A negative SAGE value indicates that a feature contributes negatively to the overall model performance, meaning that considering that feature is detrimental to the model.

\section{Evaluation \& Findings}
\label{sec:evaluation}

\begin{figure*}
    \centering
    % trim={<left> <lower> <right> <upper>}
    \includegraphics[width=0.9\textwidth, trim={11cm 13cm 9cm 12cm}]{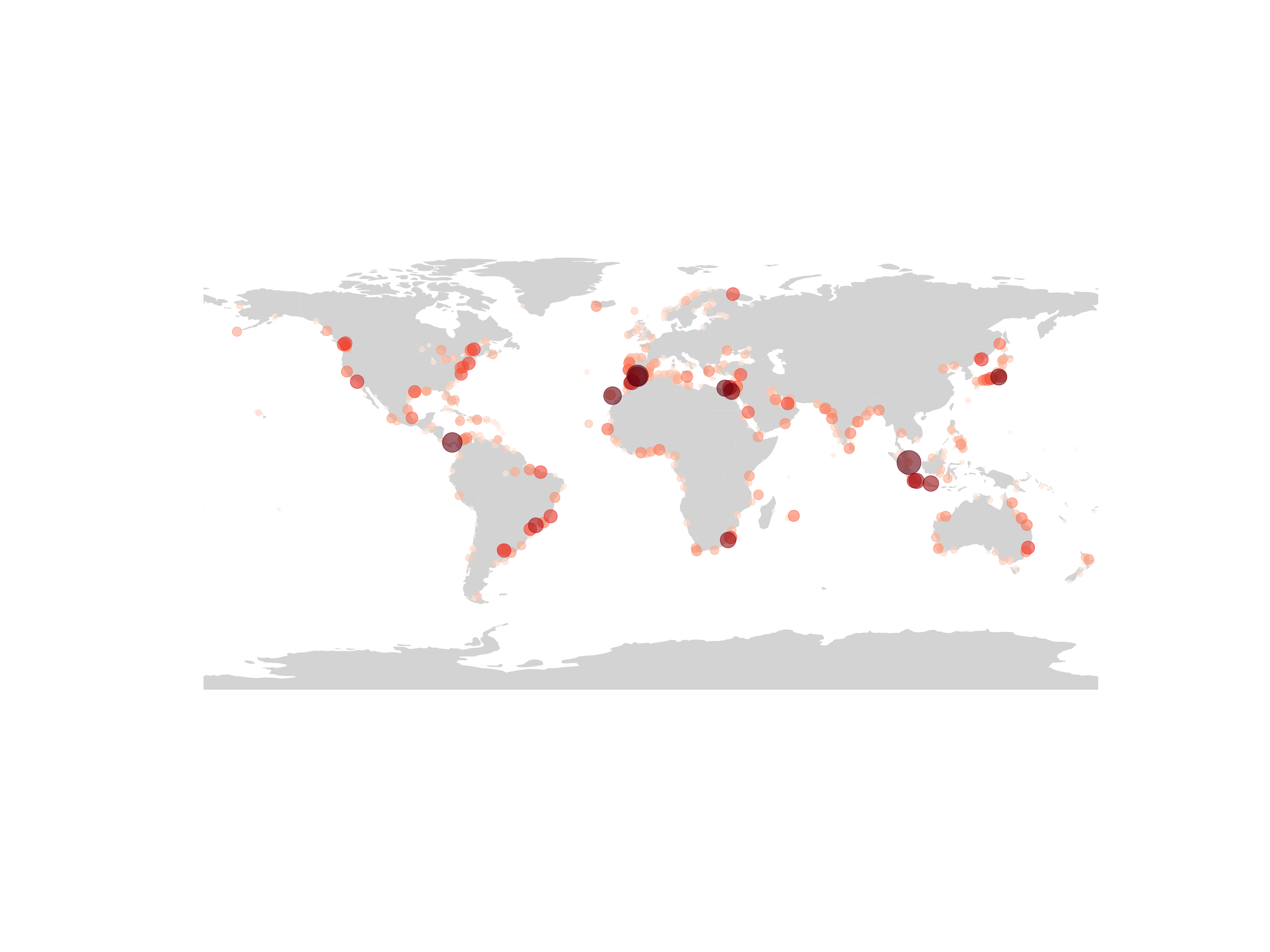}
    \captionof{figure}{Spatial Distribution and Centrality of Global Maritime Ports. This map displays the locations of major ports worldwide. The size of each circle corresponds to the importance and connectivity of the port within the global maritime network. Larger circles represent ports with higher centrality, indicating their significance in international trade and logistics.}
    \label{fig:world}
\end{figure*}

\subsection{Most central ports}
    
Figure \ref{fig:world} shows the geographical placement of the most central ports (the size and darkness of the circle are proportional to the port centrality). Similarly, Table \ref{tab:top10} shows the top 12 most central ports according to the definition of centrality given in Equation \ref{ed:centrality}. 
Examining the list in the table, we recognize some major global container hubs. The top ports show a relatively wide geographical distribution. 
First in the top 12, \textit{Keppel} in Singapore is located in one of the busiest port areas in the world.
\textit{Algeciras} in Spain and \textit{Tangier} in Morocco are located on the Strait of Gibraltar, the Mediterranean's entrance passage, and a stopping point for cargo ships. 
\textit{Puerto Cristobal} (Colon) is a ship hub on the Panama Canal's Atlantic entrance. 
\textit{Las Palmas} is the largest port of the Canary Islands, and it's strategically positioned on the Atlantic Ocean. 

\textit{Alexandria} and \textit{El-Adabiya} are two major Egyptian ports, respectively, on the Mediterranean and at the Suez Canal entrance in the Suez Gulf. 
\textit{Yokohama} is one of the largest cities in Japan, situated in a highly congested area at the entrance to Tokyo Port.
The \textit{Durban} port in South Africa is one of the largest container ports in the Southern Hemisphere.
\textit{Jakarta} and \textit{Gresik} ports are the two largest in Indonesia, with Jakarta being one of the largest in Southeast Asia. Finally, \textit{Casablanca} is the second-largest city in Morocco.\\[.25cm]

\begin{table}[h]
    \centering
    \footnotesize
    \resizebox{.48\textwidth}{!}{
    \begin{tabular}{rrlc}
    \toprule
    {} Rank & Port Name &  Country & Centrality\\
    \midrule
    1 & KEPPEL & Singapore & 10.594 \\
    2 & ALGECIRAS & Spain & 8.554 \\
    3 & PUERTO CRISTOBAL & Panama & 7.182 \\
    4 & TANGIER & Morocco & 6.742 \\
    5 & LAS PALMAS & Spain & 5.89 \\
    6 & ALEXANDRIA & Egypt & 4.941 \\
    7 & YOKOHAMA & Japan & 4.882 \\
    8 & EL-ADABIYA & Egypt & 4.823 \\
    9 & GRESIK & Indonesia & 4.542 \\
    10 & DURBAN &  South Africa & 4.538 \\
    11 & JAKARTA & Indonesia & 4.4 \\
    12 & CASABLANCA & Morocco   & 4.271 \\

\bottomrule
    \end{tabular}
    }
    \caption{Top 12 central ports and associated centrality values.}
    \label{tab:top10}
\end{table}

%This result aligns with our intuition because...
\begin{figure}[t]
    \centering
    \includegraphics[width=\columnwidth]{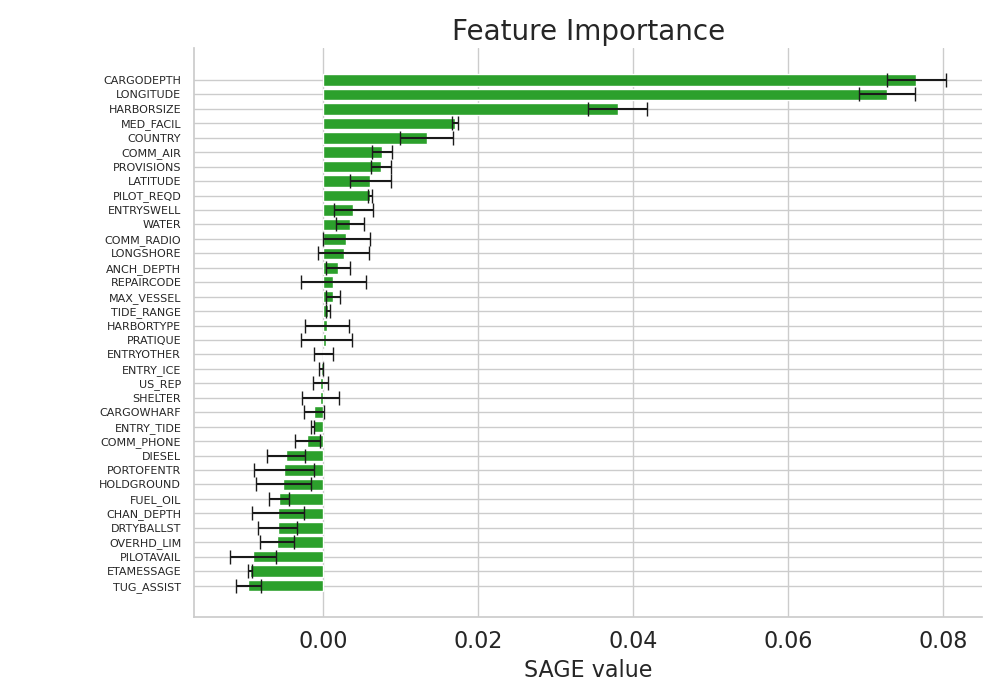}
    \caption{SAGE-Derived Feature Importance for Central Port Classification. This bar chart illustrates the global feature importance scores computed using the SAGE method for identifying the top $10\%$ most central ports. Features such as CARGODEPTH, LONGITUDE, HARBORSIZE, and MED\_FACIL exhibit the highest contributions, underscoring their relevance in the classification model. The error bars in the image denote the variability in importance across multiple runs.}
    \label{fig:sage_10_pct}
\end{figure}

\begin{figure}[!t]
    \centering
    \includegraphics[width=.98\columnwidth]{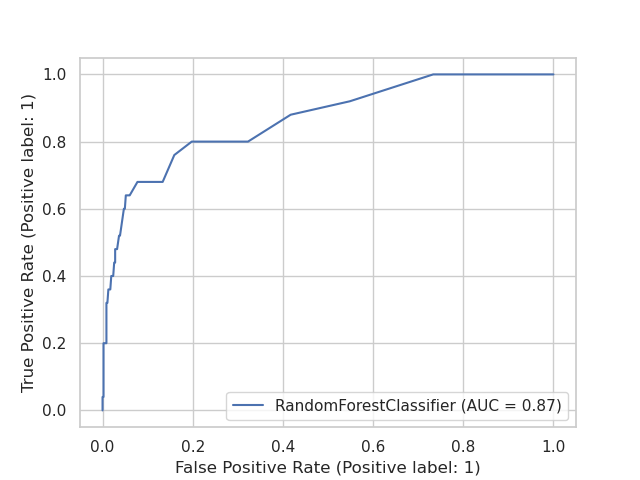}
    \includegraphics[width=.98\columnwidth]{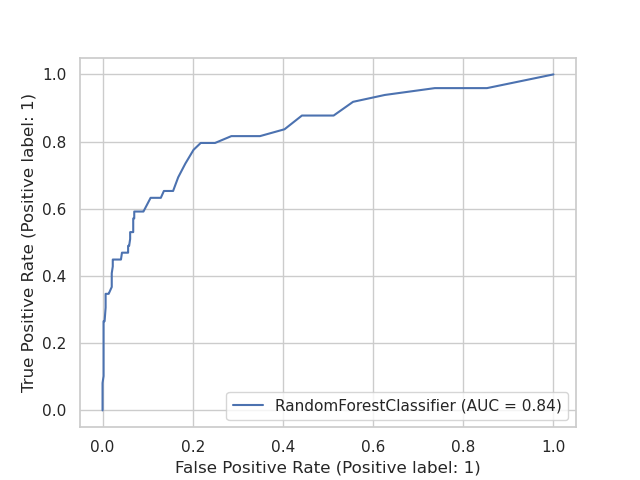}
    \includegraphics[width=.98\columnwidth]{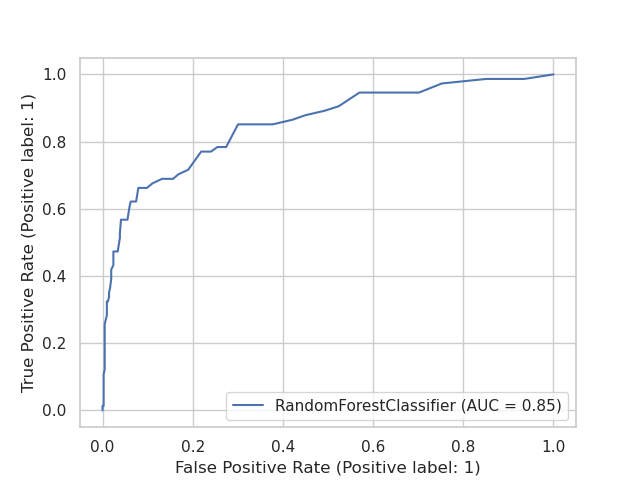}
        \caption{ROC curve from the random forest for the binary classification of most central ports. From left to right, we consider relevant ports, the ones on the $5\%$, $10\%$, $15\%$}
    \label{fig:roc_curve}
\end{figure}

\begin{figure}[!t]
    \centering
    \includegraphics[width=.98\columnwidth]{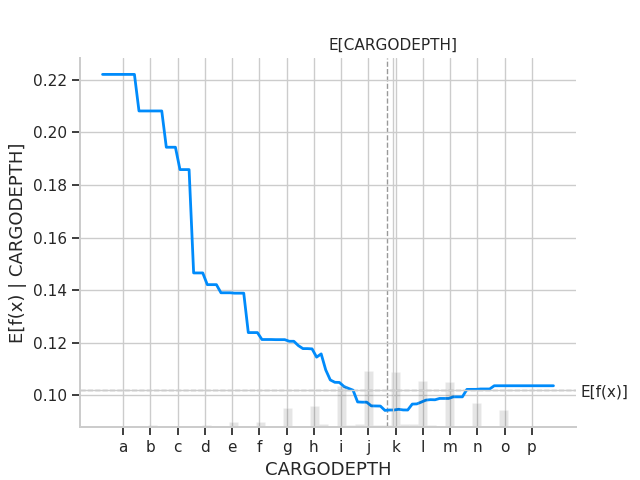}
    \includegraphics[width=.98\columnwidth]{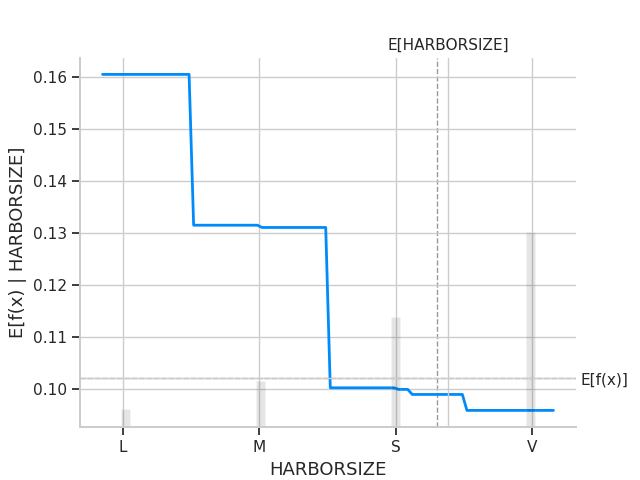}
        \caption{Partial dependence plots for the $3$ most important features according to SAGE, CARGODEPTH (top)
        %, LONGITUDE 
        and HARBOURSIZE (bottom). The vertical gray bar represents the average value of the feature. The blue partial dependence plot line is the average value of the model output when we fix the feature at hand to a given value. The grey histograms on the x-axis indicate the distribution of each feature.
        }
    \label{fig:partial_dep_plots}
\end{figure}

\subsection{Assessing Model performances}

To address the research questions we previously posed, we need to quantitatively assess how accurately the random forest model can solve the binary classification task defined in Section \ref{sec:ml}. 
To this end, we need a standard measure to assess how effectively a port is highly central in the graph of vessel voyages by examining its features alone \textbf{(RQ2)}.
Moreover, to interpret the importance of features in solving this ML task, we need to ensure that the model identifies a pattern in the data instead of returning random guesses.

The area under the \textit{Receiver Operating Characteristic} (ROC) curve is a widely used indicator of prediction accuracy for binary classification problems. 
The ROC curve is obtained by exploring various thresholds for a model's binary classification and plotting the True Positive rate versus the False Positive rate.
The resulting curve describes the performance of the binary classifier at hand, and the area under it is referred to as the Area Under the Curve (AUC).
In short, the closer the ROC is to the top left angle, the better the predictions, {\it i.e.}, AUC close to $1$.
For reference, a random classifier would be close to the diagonal line, with an AUC of $0.5$.

For all three threshold configurations for the most central ports ($5\%$, $10\%$, $15\%$), we show in Figure \ref{fig:roc_curve} the ROC curve for the binary classification of the most central ports, and the AUC in Table \ref{tab:auc_sage}.
The results confirm that the random forests obtain good results in the classification task and that the following analysis of feature importance is based on an accurate classifier, as the values of the AUC obtained are well above the $0.5$ ($0.84$, $0.88$, $0.87$, respectively) value that would be received by random classification.

% trim={<left> <wer> <right> <upper>}
\begin{figure*}[!t]
    \centering
    \includegraphics[width=\textwidth, trim={130, 0, 120, 0}]{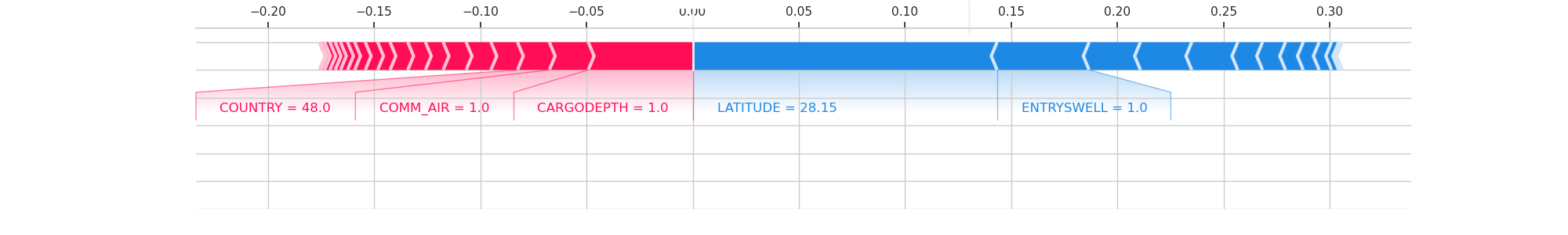}
    \caption{SHAP Force Plot for Feature Contributions at Las Palmas Port. This force plot visualizes the impact of various features on the classification outcome for the Las Palmas port case. Features contributing to a correct classification are highlighted in red, while those contributing to an incorrect classification are shown in blue. Notable features include COUNTRY, COMM\_AIR, CARGODEPTH, LATITUDE, and ENTRYSWELL, with their respective values influencing the model's prediction.}
    \label{fig:las_palmas}
\end{figure*}

\subsection{Feature Importance}

Subsequently, we analyze the results of the feature importance from the random forest model \textbf{(RQ1)}, discussing predictability and co-dependence between the aggregated centrality and port features.

\subsubsection{Global feature importance}
In Figure \ref{fig:sage_10_pct}, we show the SAGE values describing feature importance for the whole Port Network when we considered the top $10\%$ of the ports as central. It becomes clear that the most important features in predicting port centrality are CARGODEPTH, LONGITUDE, and HARBORSIZE.

\begin{table}[!b]
\vspace{.5cm}
\centering
\begin{tabular}{llll}
\midrule
Most Central Ports Threshold & 5\% &  10\% & 15\% \\
\midrule
AUC value  &    $  0.87  $ &     $  0.84  $ &     $  0.85  $ \\
\midrule
CARGODEPTH  &    $  0.02 $&   $  0.04 $&    $  0.08 $\\
LONGITUDE   &    $  0.07 $&   $  0.04 $&    $  0.07 $\\
HARBORSIZE  &    $  0.03 $&   $  0.02 $&    $  0.04 $\\
COUNTRY     &    $  0.01 $&   $  0.03 $&    $  0.01 $\\
HAN\_DEPTH &    $  0.02 $&   $  0.02 $&   $ 0.006 $\\
\bottomrule
\end{tabular}
\caption{The $5$ most important features and their SAGE values for predicting the most central ports in a binary classification}\label{tab:auc_sage}
\end{table}

\begin{table*}[!t]
    \centering
    \small
    \begin{tabular}{r|c|p{12cm}}
    \hline
        \textbf{Feature} & \textbf{Avg. Rank} & \textbf{Description} 
        \\
        \hline
        US REPRESENTATIVE & 9.8 & Indicates whether the United States maintains either civilian or military representation in the port. \\
        \hline
        DIRTY BALLAST & 12.2 & Whether a port has sufficient facilities for receiving oily or contaminated ballast. 
        \\
        \hline
        PRATIQUE & 12.6 & Whether medical practice is applied to vessels arriving in the port.
        \\
        \hline
        CARGODEPTH & 13.8 & The Greatest depth for cargo vessels available in the port.
        \\
        \hline
        CHANNEL DEPTH & 14.3 & The depth of the deepest channel leading to the port.\\
        \hline
    \end{tabular}
    \caption{Top 5 non-geographical features ordered by their average rank}
    \label{tab:local_rank}
    \vspace{-.75cm}
\end{table*}

CARGODEPTH indicates the maximum depth available for cargo vessels in the port. A greater depth allows for large vessels to visit the port. For example, cargo vessels require a depth greater than 12 meters, up to 25 meters for deeper ships.
The dataset's depth is coded using letters from A to Q, with Q representing the deepest level. Each subsequent letter increases the depth by $5$ feet, approximately $1.5$ meters. For example, the letter H corresponds to 43 feet ($\approx$13 meters).
The information about the depth is the most useful for discriminating whether a cargo ship can enter the port.

The HARBORSIZE is based on several factors, including area, facilities, and wharf space. It is codified into four categories: Large (L), Medium (M), Small (S), or Very Small (V).
Similar to cargo depth, most cargo visits are focused on large ports, which makes this feature one of the most informative.
Finally, LONGITUDE shows the importance of a port's geographical position. It is interesting to note how longitude is more relevant than latitude in this context. One possible explanation is that most central ports, as graphically illustrated in Figure \ref{fig:world}, are located in specific longitudinal areas, including the Americas, Europe, and Southeast Asia. Longitude increases accessibility and is a good discriminator for central ports.

While SAGE quantifies the importance of a feature in improving the model's prediction, it does not provide information on the effect of different values of each variable on the model's prediction. 
To gain additional insight, Figure \ref{fig:partial_dep_plots} shows the partial dependency plots for the three features that were globally most important. These plots illustrate how the model's prediction changes on average when the value of one variable is altered. Notice that the dependencies are highly non-linear due to the non-linear nature of the model we employed. Moreover, the CARGODEPTH data means higher values correspond to lower values in meters. Hence, increasing the actual depth increases the likelihood that a port will be predicted as highly central. Finally, the correlation between a harbor size (HARBORSIZE) and port centrality is made evident in the plot.

\subsubsection{Local Feature Importance}

Local feature importance measures the relevance of a feature in classifying single-input items.
To summarize what features are more locally important, we ranked them according to their average contribution to each port's classification.
Intuitively, the average ranking would measure how consistently a feature contributes to a port's positive classification.
In general, the geographical features ranked the highest. Apart from those, Table \ref{tab:local_rank} shows the top 5 non-geographical features. Interestingly, a ``political" feature ranks first. Then, services such as medical procedures and management of contaminated materials are typical of international ports and could differentiate the port centrality. Finally, port depths directly correlate with the port's ability to accommodate large cargo ships, a further centrality indicator.

Another interesting analysis involves examining individual ports and how their features correlate with their local importance. Figure \ref{fig:las_palmas} shows the Las Palmas port. CARGODEPTH, COMM\_AIR (indicates whether airport communications are available), and COUNTRY are features that help classify them. Contrarily, its latitude and ENTRY SWELL (binary, whether there is a natural factor restricting the entrance of vessels) are not favorable features for this port.

\section{Conclusion}
\label{sec:conclusion}

We use explainable machine learning to identify key features of significant ports.
To achieve this, we constructed a Ports Network using three years of worldwide data, where the significance of a port is determined by the combination of centrality measures commonly used in the literature.
We performed a machine learning task to predict the port's importance using publicly available features and analyzed which features are most useful for inference.

Geographical features are the most informative for identifying central ports, {\it i.e.}, a direct and expected outcome, given that ports are inherently tied to specific coastal locations. Beyond geography, features related to port depth, such as entrance and pier depth, also play a critical role in accurate classification. This aligns with expectations, as the analysis focuses on cargo vessels, which typically require deep-water infrastructure for safe access and docking.

This paper also paves the way for similar research on other types of vessel modalities, such as passenger or leisure vessels.
Additionally, identifying the features of central maritime ports can facilitate research on developing new or existing ports and provide a tool for studying inter-port and regional relationships.
These points provide potential research directions for extending this study.

\section*{Acknowledgement}
The authors acknowledge the support of the H2020 EU Project MASTER (Multiple ASpects TrajEctoRy management and analysis), funded under the Marie Skłodowska-Curie grant agreement No. 777695. This research was partially supported by the Institute for Big Data Analytics (IBDA) and the Ocean Frontier Institute (OFI) at Dalhousie University, Halifax, NS, Canada, as well as by the Canadian Foundation for Innovation’s MERIDIAN Cyberinfrastructure. Additional funding was provided by the \textit{Conselho Nacional de Desenvolvimento Científico e Tecnológico} (CNPq), Brazil.

% \section*{Declaration of competing interest}
% \noindent%
% The authors declare no conflicts of interest.

\section*{Data Licensing and Disclosure}
The research described in this paper used data acquired from Spire under MERIDIAN's fair-use and non-disclosure data license. Due to licensing agreements, we are unable to share the raw data. However, we can provide a trained model and the means to retrain the models using open-source data from similar maritime regions.

\balance  % Balance the columns
\bibliographystyle{unsrtnat}
\bibliography{biblio}

\end{document}